\documentclass[journal]{IEEEtran}
\usepackage{subfigure}
\usepackage[dvips]{graphicx}
\usepackage[english]{babel}
\usepackage[latin1]{inputenc}
\usepackage{amsmath} 
\usepackage{amssymb}
\usepackage{url}
\usepackage{color}
\usepackage{cite} 
\usepackage{stfloats}  

\ifCLASSINFOpdf
\else
\fi
\begin{document}
\onecolumn

%
\title{On Kenn's Rule of Combination\\
Applied to Breast Cancer Precision Therapy}


\markboth{Technical Note - TN-2023-02-28, February 2023.}{}

\author{Jean~Dezert, and~Albena~Tchamova
\thanks{J. Dezert is with the Information Processing and Systems Dept. at The French Aerospace Lab, Palaiseau 91120, France.  e-mail: jean.dezert@onera.fr.}
\thanks{A. Tchamova is with the IICT, Bulgarian Academy of Sciences, Sofia, Bulgaria.  e-mail: tchamova@bas.bg.}
\thanks{}
\thanks{{\bf{Citation}}: J.~Dezert, A. Tchamova, On Kenn's Rule of Combination Applied to Breast Cancer Precision Therapy, Technical Note TN-2023-02-28, Feb. 2023.}
}

\maketitle

\begin{abstract}
This short technical note points out an erroneous claim about a new rule of combination of basic belief assignments presented recently by Kenn et al. in \cite{Kenn2023}, referred as Kenn's rule of combination (or just as KRC for short). We prove thanks a very simple counter-example that Kenn's rule is not associative. Consequently, the results of the method proposed by Kenn et al. highly depends on the ad-hoc sequential order chosen for the fusion process as proposed by the authors. This serious problem casts in doubt the interest of this method and its real ability to provide trustful results and to make good decisions to help for precise breast cancer therapy.
\end{abstract}


\vspace{1cm}

\IEEEpeerreviewmaketitle

\section{Introduction}

Recently a paper devoted to the Breast Cancer Precision Therapy by Kenn et al. \cite{Kenn2023} attracted our attention for two main reasons: 1) this application of information fusion is very interesting and important; 2) Kenn's et al. method is based on a new rule of combination of basic belief assignments (BBAs). Because we did some theoretical contributions in this field \cite{WebJean} we wanted to examine this new combination rule in detail. So, we have read with interest Kenn's et al paper. Unfortunately we quickly discovered a serious erroneous claim about Kenn's rule of combination (KRC) and this has strong consequences on the methodology presented by Kenn. In this short technical note we warn the readers of the risk of potential therapy errors if such a method is used in practice. We clearly explain the problem of the method presented by Kenn et al. To make the paper self-containing, we recall briefly the basics of belief functions in the next section, and the KRC in the section \ref{KRCsec}. In section \ref{SecCounterExample} we prove thanks a very simple numerical counter-example that KRC is not associative. Conclusion and recommendations are given in the last section of this note.

\section{Belief functions}
\label{BFsec}

The belief functions (BF) were introduced by Shafer \cite{Shafer1976} for modeling epistemic uncertainty, reasoning about uncertainty and combining distinct sources of evidence. The answer of the problem under concern is assumed to belong to a known finite discrete frame of discernement (FoD) $\Theta=\{\theta_1,\ldots,\theta_N\}$ where all elements (i.e. members) of $\Theta$ are exhaustive and exclusive. The set of all subsets of $\Theta$ (including empty set $\emptyset$, and $\Theta$) is the power-set of $\Theta$ denoted by $2^\Theta$. The number of elements (i.e. the cardinality) of the power-set is $2^{|\Theta |}$. A (normalized) basic belief assignment (BBA) associated with a given source of evidence is a mapping $m^\Theta(\cdot):2^\Theta \to [0,1]$ such that $m^\Theta(\emptyset)=0$ and $\sum_{X\in 2^\Theta} m^\Theta(X) = 1$. A BBA $m^\Theta(\cdot)$ characterizes a source of evidence related with a FoD $\Theta$. For notation shorthand, we can omit the superscript $\Theta$ in $m^\Theta(\cdot)$ notation if there is no ambiguity on the FoD we work with. The quantity $m(X)$ is called the mass of belief of $X$. The element $X\in 2^\Theta$ is called a focal element (FE) of $m(\cdot)$ if $m(X)>0$. The set of all focal elements of $m(\cdot)$ is denoted\footnote{$\triangleq$ means {\it{equal by definition}}.} by $\mathcal{F}_{\Theta}(m)\triangleq \{X\in 2^\Theta | m(X)>0 \}$. The belief and the plausibility of $X$ are respectively defined for any $X\in 2^\Theta$ by \cite{Shafer1976}
\begin{equation}
Bel(X) = \sum_{Y\in 2^\Theta | Y\subseteq X} m(Y)
\label{eqBel}
\end{equation}
\begin{equation}
Pl(X) =  \sum_{Y\in 2^\Theta | X\cap Y\neq \emptyset} m(Y)=1-\text{Bel}(\bar{X}).
\label{eqPl}
\end{equation}
\noindent
where ${\bar{X}\triangleq\Theta\setminus\{X\}}$ is the complement of $X$ in $\Theta$.

One has always ${0\leq Bel(X)\leq Pl(X) \leq 1}$, see \cite{Shafer1976}. For ${X=\emptyset}$, ${Bel(\emptyset)=Pl(\emptyset)=0}$, and for ${X=\Theta}$ one has ${Bel(\Theta)=Pl(\Theta)=1}$. $Bel(X)$ and $Pl(X)$ are often interpreted as the lower and upper bounds of unknown probability $P(X)$ of $X$, that is ${Bel(X) \leq P(X) \leq Pl(X)}$. To quantify the uncertainty (i.e. the imprecision) of ${P(X)\in [Bel(X), Pl(X)]}$, we use ${u(X)\in [0,1]}$ defined by
\begin{equation}
u(X)\triangleq Pl(X)-Bel(X)
\label{defuX}
\end{equation}
If ${u(X)=0}$, ${Bel(X)=Pl(X)}$ and therefore $P(X)$ is known precisely because ${P(X)=Bel(X)=Pl(X)}$.
One has ${u(\emptyset)=0}$ because ${Bel(\emptyset)=Pl(\emptyset)=0}$, and one has ${u(\Theta)=0}$ because ${Bel(\Theta)=Pl(\Theta)=1}$. If all focal elements of $m(\cdot)$ are singletons of $2^\Theta$ the BBA $m(\cdot)$ is a Bayesian BBA because ${\forall X\in2^\Theta}$ one has ${Bel(X)=Pl(X)=P(X)}$ and $u(X)=0$. Hence the belief and plausibility of $X$ coincide with a probability measure $P(X)$ defined on the FoD $\Theta$. The vacuous BBA characterizing a totally ignorant source of evidence is defined by ${m_v(X)=1}$ for ${X=\Theta}$, and ${m_v(X)=0}$ for all ${X\in 2^\Theta}$ different of $\Theta$.

In the Mathematical Theory of Evidence of Glenn Shafer, the combination of two BBAs $m_1(.)$ and $m_2(.)$ defined over the same FoD $\Theta$ is obtained with Dempster's rule. More precisely\footnote{DS upper index in formula \eqref{DSrule} stands for Dempster-Shafer because this rule is often referred also as Dempster-Shafer rule of combination in the literature.} by $m_{12}^{DS}(\emptyset)=0$, and for any $X\in 2^\Theta\setminus\{\emptyset\}$ by
\begin{equation}
m_{12}^{DS}(X)=\frac{\sum_{\substack{X_1,X_2\subseteq \Theta \\ X_1\cap X_2=X}} m_1(X_1)m_2(X_2)}{1-\sum_{\substack{X_1,X_2\subseteq \Theta \\ X_1\cap X_2=\emptyset}}m_1(X_1)m_2(X_2)}
\label{DSrule}
\end{equation}

The value $K_{12}=\sum_{\substack{X_1,X_2 \subseteq \Theta \\ X_1\cap X_2=\emptyset}}m_1(X_1)m_2(X_2)$ is classically interpreted as the degree of conflict between the BBAs $m_1(.)$ and $m_2(.)$.
When the degree of conflict is maximum one has $K_{12}=1$, and in this particular case Dempster-Shafer rule cannot be applied because of division by zero in the formula \eqref{DSrule}. This rule can be easily directly generalized for the combination of more than two BBAs.

The DS rule has had a great success during the past decades in expert systems and artificial intelligence mainly because it is a commutative and associative rule of combination able to deal with (possibly epistemic) uncertainty and incomplete information based on an appealing mathematical framework. This makes its use very attractive from the implementation standpoint in decision-making support systems. Indeed, because of its associativity we can apply DS rule sequentially when we have more than two sources of evidence to fuse, and the sequence order will not impact the DS fusion result. Unfortunately, DS rule of combination generates counter-intuitive results due to the normalization step in DS formula (not only in high conflicting situations but also in low conflicting situations as well), and it generates very controversial dictatorial behaviors, see \cite{Zadeh1986,Fusion2012} for discussions with examples. That is why many alternatives of DS rule have been proposed in the literature to circumvent these serious problems of DS rule. Unfortunately, there is no general consensus in the scientific community about the choice of the {\it{best}} rule of combination of belief functions to make for the applications.

\section{Kenn's rule of combination}
\label{KRCsec}

Kenn's rule of combination (KRC) proposed in \cite{Kenn2023} is a slight modification of DS rule introducing a tuning parameter $\lambda\in [0,1]$. The KRC of two BBAs $m_1(.)$ and $m_2(.)$ defined over the same FoD $\Theta$ is denoted symbolically $m_1\oplus_\lambda m_2$ in \cite{Kenn2023}. Its mathematical expression is given by\footnote{see formula (4) in \cite{Kenn2023}} $m_{12}^{KRC}(\emptyset)=0$, for $X\in 2^{\Theta}\setminus\{\emptyset\}$ by

\begin{equation}
m_{12}^{KRC}(X)=[m_1\oplus_\lambda m_2](X)=\frac{\sum_{\substack{X_1,X_2\subseteq \Theta \\ X_1\cap X_2=X}} m_1(X_1)m_2(X_2)}{1-\lambda \sum_{\substack{X_1,X_2\subseteq \Theta \\ X_1\cap X_2=\emptyset}}m_1(X_1)m_2(X_2)}
\label{KRCrule1}
\end{equation}
and for ${X=\Theta}$ by
\begin{equation}
m_{12}^{KRC}(\Theta)=[m_1\oplus_\lambda m_2](\Theta)=1-\sum_{X\subset \Theta} m_{12}^{KRC}(X)
\label{KRCrule2}
\end{equation}

For $\lambda=1$, KRC coincides with Dempster-Shafer rule and consequently it will suffer of same problems as DS rule in this particular case. According to Kenn et al., the parameter $\lambda$ in the formula \eqref{KRCrule1} provides flexibility to adapt to circumstances and the
restriction to $\lambda \leq 1$ is motivated by restricting the authors to an interpolation type evidential combination rule.
KRC is claimed associative and commutative by Kenn et al. (see page 5 of  \cite{Kenn2023}). We prove in the next section that KRC is in fact not associative. Because of non-associativity of KRC, the methodology proposed in \cite{Kenn2023} becomes very disputable and doubtful, and potentially very dangerous for breast cancer therapy application addressed by Kenn et al., and for any other applications using sequential fusion of sources of evidences based on KRC.

\section{Counter-example of associativity of KRC}
\label{SecCounterExample}

To prove that KRC is not associative it suffices to verify that
\begin{equation}
(m_1\oplus_\lambda m_2)\oplus_\lambda m_3 \neq m_1\oplus_\lambda (m_2 \oplus_\lambda m_3)
\label{NonAssoc}
\end{equation}

To prove \eqref{NonAssoc} when $\lambda < 1$, just consider for instance $\lambda=0.2$, the FoD $\Theta=\{A,B\}$ and the three BBAs given by
$$m_1(A)=0.2, m_1(B)=0.7, m_1(A\cup B)=0.1$$
$$m_2(A)=0.8, m_2(B)=0.1, m_2(A\cup B)=0.1$$
$$m_3(A)=0.4, m_3(B)=0.3, m_3(A\cup B)=0.3$$

\subsection{Derivation of $(m_1 \oplus_{0.2} m_2)\oplus_{0.2} m_3$ }

For the combination of $m_1$ with $m_2$ we have the degree of conflict
$$K_{12}=m_1(A)m_2(B)+m_1(B)m_2(A)=0.2\cdot 0.1 + 0.7\cdot 0.8=0.58$$
The results of the conjunctive fusion of $m_1$ with $m_2$ for $A$ and $B$ are
\begin{align*}
m_{12}(A)& =m_1(A)m_2(A)+m_1(A)m_2(A\cup B)+m_2(A)m_1(A\cup B)\\
& = 0.2\cdot 0.8 + 0.2\cdot 0.1 + 0.8\cdot 0.1 = 0.26\\
m_{12}(B)& =m_1(B)m_2(B)+m_1(B)m_2(A\cup B)+m_2(B)m_1(A\cup B)\\
& = 0.7\cdot 0.1 + 0.7\cdot 0.1 + 0.1\cdot 0.1 = 0.15
\end{align*}
For KRC of $m_1$ with $m_2$ we get (taking $\lambda=0.2$) $m_{12}^{KRC}(\emptyset)=0$ and
$$m_{12}^{KRC}(A)=[m_1 \oplus_{\lambda} m_2](A)=\frac{m_{12}(A)}{1-\lambda K_{12}}= \frac{0.26}{1-0.2\cdot 0.58}\approx 0.2941$$
$$m_{12}^{KRC}(B)=[m_1 \oplus_{\lambda} m_2](B)=\frac{m_{12}(B)}{1-\lambda K_{12}}= \frac{0.15}{1-0.2\cdot 0.58}\approx 0.1697$$
$$m_{12}^{KRC}(A\cup B)=[m_1 \oplus_{\lambda} m_2](A\cup B)=1 -  \frac{0.26}{1-0.2\cdot 0.58} - \frac{0.15}{1-0.2\cdot 0.58} \approx 0.5362 $$

For the combination of $m_{12}^{KRC}=m_1\oplus_\lambda m_2$ with $m_3$ we have the degree of conflict
$$K_{(12)3}=m_{12}^{KRC}(A)m_3(B)+m_{12}^{KRC}(B)m_3(A)\approx 0.2941\cdot 0.3 + 0.1697 \cdot 0.4\approx 0.1561$$

The results of the conjunctive fusion of $m_{12}^{KRC}$ with $m_3$ for $A$ and $B$ are
\begin{align*}
m_{(12)3}(A)& =m_{12}^{KRC}(A)m_3(A)+m_{12}^{KRC}(A)m_3(A\cup B)+m_3(A)m_{12}^{KRC}(A\cup B)\\
& \approx 0.2941 \cdot 0.4 + 0.2941\cdot 0.3 + 0.4\cdot 0.5362 = 0.4204
\end{align*}
\begin{align*}
m_{(12)3}(B)& =m_{12}^{KRC}(B)m_3(B)+m_{12}^{KRC}(B)m_3(A\cup B)+m_3(B)m_{12}^{KRC}(A\cup B)\\
& \approx 0.1697\cdot 0.3 + 0.1697\cdot 0.3 + 0.3\cdot 0.5362 = 0.2627
\end{align*}

Therefore, the KRC of $m_{12}^{KRC}$ with $m_3$ yields $m_{(12)3}^{KRC}(\emptyset)=0$ and
$$m_{(12)3}^{KRC}(A)=[m_{12}^{KRC}\oplus_\lambda m_3](A)=\frac{m_{(12)3}(A)}{1-\lambda K_{(12)3}}= \frac{0.4204}{1-0.2\cdot 0.1561}\approx 0.4339 $$
$$m_{(12)3}^{KRC}(B)=[m_{12}^{KRC}\oplus_\lambda m_3](B)=\frac{m_{(12)3}(B)}{1-\lambda K_{(12)3}}= \frac{0.2627}{1-0.2\cdot 0.1561}\approx 0.2711 $$
$$m_{(12)3}^{KRC}(A\cup B)=[m_{12}^{KRC}\oplus_\lambda m_3](A\cup B)\approx 1 -  0.4339  - 0.2711 \approx 0.2950$$
Hence for the fusion $(m_1 \oplus_{0.2} m_2)\oplus_{0.2} m_3$ we get finally
\begin{align}
m_{(12)3}^{KRC}(A)& =[(m_1 \oplus_{0.2} m_2)\oplus_{0.2} m_3](A)\approx 0.4339\\
m_{(12)3}^{KRC}(B)& =[(m_1 \oplus_{0.2} m_2)\oplus_{0.2} m_3](B)\approx 0.2711\\
m_{(12)3}^{KRC}(A\cup B)& =[(m_1 \oplus_{0.2} m_2)\oplus_{0.2} m_3](A\cup B)\approx 0.2950
\end{align}

\subsection{Derivation of $m_1 \oplus_{0.2} (m_2\oplus_{0.2} m_3)$ }

For the combination of $m_2$ with $m_3$ we have
$$K_{23}=m_2(A)m_3(B)+m_2(B)m_3(A)=0.8\cdot 0.3 + 0.1\cdot 0.4=0.28$$
The results of the conjunctive fusion of $m_2$ with $m_3$ for $A$ and $B$ are
\begin{align*}
m_{23}(A)& =m_2(A)m_3(A)+m_2(A)m_3(A\cup B)+m_3(A)m_2(A\cup B)\\
& = 0.8\cdot 0.4 + 0.8\cdot 0.3 + 0.4\cdot 0.1 = 0.60
\end{align*}
\begin{align*}
m_{23}(B)& =m_2(B)m_3(B)+m_2(B)m_3(A\cup B)+m_3(B)m_2(A\cup B)\\
& = 0.1\cdot 0.3 + 0.1\cdot 0.3 + 0.3\cdot 0.1 = 0.09
\end{align*}

For KRC of $m_2$ with $m_3$ we get (taking $\lambda=0.2$) $m_{23}^{KRC}(\emptyset)=0$ and
$$m_{23}^{KRC}(A)=[m_2 \oplus_{\lambda} m_3](A)=\frac{m_{23}(A)}{1-\lambda K_{23}}= \frac{0.60}{1-0.2\cdot 0.28}\approx 0.6356$$
$$m_{23}^{KRC}(B)=[m_2 \oplus_{\lambda} m_3](B)=\frac{m_{23}(B)}{1-\lambda K_{23}}= \frac{ 0.09}{1-0.2\cdot 0.28}\approx 0.0953$$
$$m_{23}^{KRC}(A\cup B)=[m_2 \oplus_{\lambda} m_3](A\cup B)\approx 1 -  0.6356 - 0.0953 \approx 0.2691 $$

For the combination of $m_1$ with $m_{23}^{KRC}=m_2\oplus_\lambda m_3$  we have the degree of conflict
$$K_{1(23)}=m_{23}^{KRC}(A)m_1(B)+m_{23}^{KRC}(B)m_1(A)\approx 0.63563\cdot 0.7 + 0.0953 \cdot 0.2 \approx 0.4640 $$

The results of the conjunctive fusion of $m_1$ with $m_{23}^{KRC}$ for $A$ and $B$ are
\begin{align*}
m_{1(23)}(A)& =m_{23}^{KRC}(A)m_1(A)+m_{23}^{KRC}(A)m_1(A\cup B)+m_1(A)m_{23}^{KRC}(A\cup B)\\
& \approx  0.6356 \cdot 0.2 + 0.6356 \cdot 0.1 + 0.2\cdot 0.2691 \approx 0.2445
\end{align*}
\begin{align*}
m_{1(23)}(B)& =m_{23}^{KRC}(B)m_1(B)+m_{23}^{KRC}(B)m_1(A\cup B)+m_1(B)m_{23}^{KRC}(A\cup B)\\
& \approx 0.0953\cdot 0.7 +0.0953\cdot 0.1 + 0.7\cdot 0.2691 \approx 0.2646
\end{align*}

Therefore, the KRC of $m_1$ with $m_{23}^{KRC}$ yields $m_{1(23)}^{KRC}(\emptyset)=0$ and
$$m_{1(23)}^{KRC}(A)=[m_1 \oplus_\lambda m_{23}^{KRC}](A)=\frac{m_{1(23)}(A)}{1-\lambda K_{1(23)}}= \frac{0.2445}{1-0.2\cdot 0.4640}\approx 0.2695 $$
$$m_{1(23)}^{KRC}(B)=[m_1 \oplus_\lambda m_{23}^{KRC}](B)=\frac{m_{1(23)}(B)}{1-\lambda K_{1(23)}}= \frac{0.2646}{1-0.2\cdot 0.4640}\approx 0.2917 $$
$$m_{1(23)}^{KRC}(A\cup B)=[m_1 \oplus_\lambda m_{23}^{KRC}](A\cup B)\approx 1 -  0.2695  - 0.2917 \approx 0.4388$$
Hence for the fusion $m_1 \oplus_{0.2} (m_2 \oplus_{0.2} m_3)$ we get finally
\begin{align}
m_{1(23)}^{KRC}(A)& =[m_1 \oplus_{0.2} (m_2 \oplus_{0.2} m_3)](A)\approx 0.2695\\
m_{1(23)}^{KRC}(B)& =[m_1 \oplus_{0.2} (m_2 \oplus_{0.2} m_3)](B)\approx 0.2917\\
m_{1(23)}^{KRC}(A\cup B)& =[m_1 \oplus_{0.2} (m_2 \oplus_{0.2} m_3)](A\cup B)\approx 0.4388
\end{align}

We see clearly that KRC is not associative because $(m_1\oplus_\lambda m_2)\oplus_\lambda m_3 \neq m_1\oplus_\lambda (m_2 \oplus_\lambda m_3)$ as reported in the Table \ref{Table1}.

{\small{
\begin{table}[htp]
\begin{center}
\begin{tabular}{|c||c|c|c||c|c|}
\hline
Elements & $m_1$ & $m_2$ & $m_3$ & $(m_1\oplus_\lambda m_2)\oplus_\lambda m_3$ & $m_1\oplus_\lambda (m_2 \oplus_\lambda m_3)$ \\
\hline
$\emptyset$ & 0   &  0   & 0   & 0         & 0 \\
$A$             & 0.2  &  0.8  & 0.4 &   0.4339   & 0.2695\\
$B$             & 0.7  &  0.1 & 0.3 &  0.2711    & 0.2917 \\
$A\cup B$  & 0.1   & 0.1 & 0.3 &   0.2950    & 0.4388 \\
\hline
\end{tabular}
\end{center}
\caption{Counter-example of associativity of KRC with $\lambda=0.2$.}
\label{Table1}
\end{table}
}}

\subsection{Comment on decision-making method used by Kenn et al. }

For our simple example we get with the sequential fusion $(m_1\oplus_{0.2} m_2)\oplus_{0.2} m_3$ the following belief intervals
\begin{align*}
& [Bel_{(12)3}(\emptyset),Pl_{(12)3}(\emptyset)]=[0,0]\\
& [Bel_{(12)3}(A),Pl_{(12)3}(A)]=[0.4339,0.7289]\\
& [Bel_{(12)3}(B),Pl_{(12)3}(B)]=[0.2711,0.5661]\\
& [Bel_{(12)3}(A\cup B),Pl_{(12)3}(A\cup B)]=[1,1]
\end{align*}
and with the sequential fusion $m_1\oplus_{0.2} (m_2\oplus_{0.2} m_3)$
\begin{align*}
& [Bel_{1(23)}(\emptyset),Pl_{1(23)}(\emptyset)]=[0,0]\\
& [Bel_{1(23)}(A),Pl_{1(23)}(A)]=[0.2695,0.7083]\\
& [Bel_{1(23)}(B),Pl_{1(23)}(B)]=[0.2917,0.7305]\\
& [Bel_{1(23)}(A\cup B),Pl_{1(23)}(A\cup B)]=[1,1]
\end{align*}

Based on these results and the decision-making method used by Kenn et al. (see section 3.4 of \cite{Kenn2023}) it is clear that no decision for $A$ or for $B$ can be made using the sequential fusion $(m_1\oplus_{0.2} m_2)\oplus_{0.2} m_3$ because we have neither $Bel_{(12)3}(A)> Pl_{(12)3}(B)$, nor $Bel_{(12)3}(B)> Pl_{(12)3}(A)$. Similarly, no decision can be drawn  for $A$ or for $B$ from the result of the sequential fusion $m_1\oplus_{0.2} (m_2\oplus_{0.2} m_3)$ because we have neither $Bel_{1(23)}(A)> Pl_{1(23)}(B)$, nor $Bel_{1(23)}(B)> Pl_{1(23)}(A)$. In fact we just could always take as final decision based on Kenn's decision-making method the whole frame of discernment because one always has $(Bel_{(12)3}(A\cup B)=1) > (Pl_{(12)3}(\emptyset)=0)$  and $(Bel_{1(23)}(A\cup B)=1) > (Pl_{1(23)}(\emptyset)=0)$ but such type of decision is obviously not useful at all for the applications because it does not help to make a clear choice between $A$ and $B$. So, the decision-making method used by Kenn et al. does not work for all cases of BBA distributions as shown in this very simple example, and that is why it is not judicious and recommended to use it.

\section{Conclusion}

The consequence of non-associativity of the method presented by Kenn et al. in \cite{Kenn2023} can have a strong impact on the results and on decision-making in general if the KRC is applied sequentially for information fusion as it is proposed by the authors in their paper (see formula (9) page 7 of \cite{Kenn2023}). Because of this problem, we have a serious concern about the interest and the effectiveness of the method presented by Kenn et al.. We warn the potential users of this approach about the high risk of wrong decisions (when they are possible which is not always the case as shown in our counter-example) based on this method. This could have dramatical therapy consequences. If the authors want to use this KRC-based approach we think they should at least better consider a global information fusion processing than a sequential one, and they should adopt a better decision-making strategy. They also should compare their results with other advanced rules of combination and use the same decision strategy to make comparisons to show the real advantages of this approach, if any. The measure of the performances of the method with real open data sets for breast cancer therapy application and ground truth is also recommended.


\begin{thebibliography}{99}

\bibitem{Kenn2023}
M.~Kenn, R.~Karch, C.F~Singer, G.~Dorffner, W.~Schreiner, \emph{Flexible Risk Evidence Combination Rules in Breast Cancer Precision Therapy}, J. Pers. Med. 2023, 13, 119.
\url{https://doi.org/10.3390/jpm13010119}

\bibitem{WebJean}
\url{https://www.onera.fr/fr/staff/jean-dezert/references}

\bibitem{Shafer1976}
G.~Shafer, \emph{A mathematical theory of evidence}, Princeton University Press, 1976.

\bibitem{Zadeh1986}
L.A.~Zadeh, \emph{A simple view of the Dempster-Shafer theory of evidence and its implication for the rule of combination}, The Al Magazine, Vol. 7 (2), pp. 85--90, 1986.


\bibitem{Fusion2012}
J.~Dezert, P.~Wang, A.~Tchamova, \emph{On The Validity of Dempster-Shafer Theory}, in Proc. of  Fusion 2012, Singapore, July 2012.

\end{thebibliography}
\end{document}